\documentclass[final]{cvpr}
\usepackage[table]{xcolor}
\usepackage{times}
\usepackage{epsfig}
\usepackage{graphicx}
\usepackage{amsmath}
\usepackage{amssymb}
\makeatletter
\@namedef{ver@everyshi.sty}{}
\makeatother
\usepackage{pgfplots}
\usepackage{pgfplotstable}
\usepackage{csvsimple}
\usetikzlibrary{positioning}
\usepackage[T1]{fontenc}

\usepackage[round-mode=places, round-integer-to-decimal, round-precision=1,
			table-number-alignment=center,
			round-integer-to-decimal]{siunitx}

\usepackage[pagebackref=true,breaklinks=true,colorlinks,bookmarks=false]{hyperref}

\def\cvprPaperID{****} 
\def\confYear{CVPR 2021}

\begin{document}

\title{Online and Real-Time Tracking in a Surveillance Scenario}

\author{
	Oliver~Urbann\(^{1,*}\), Oliver~Bredtmann\(^2\), Maximilian~Otten\(^1\), \\
	Jan-Philip Richter\(^1\), Thilo Bauer\(^2\), David~Zibriczky\(^2\)\\
	\(^{1}\)Fraunhofer IML, Dortmund, Germany \\
	\(^{2}\)DB Schenker, Essen, Germany\\
	\(^*\)\texttt{oliver.urbann@iml.fraunhofer.de}
}


\maketitle

\begin{abstract}
	This paper presents an approach for tracking in a surveillance scenario. 
Typical aspects for this scenario are a 24/7 operation with a static camera mounted above the height of a human with many objects or people. The Multiple Object Tracking Benchmark 20 (MOT20) reflects this scenario best. We can show that our approach is real-time capable on this benchmark and outperforms all other real-time capable approaches in HOTA, MOTA, and IDF1. We achieve this by contributing a fast Siamese network reformulated for linear runtime (instead of quadratic) to generate fingerprints from detections. Thus, it is possible to associate the detections to Kalman filters based on multiple tracking specific ratings: Cosine similarity of fingerprints, Intersection over Union, and pixel distance ratio in the image.
\end{abstract}

\section{Introduction}

Tracking is a broad research area with a long history and a wide area of application. This paper focuses on scenarios in a typical surveillance application: A 24/7 video stream where many objects or persons must be tracked at the same time. Here, cameras are usually mounted at a height that reduces occlusions and have fixed positions and angles. Due to the 24/7 operation, the tracking algorithm must run in real-time to avoid a growing buffer with unprocessed data. Typical applications are in warehouses optimizing material routing or fork lifter paths, passenger routing in airports to reduce queues, or crowd management in a sports stadium.
The MOT20 dataset \cite{MOTChallenge20} reflects all these challenges best and is thus chosen for evaluation in this paper. It furthermore includes day and night scenes and provides a frame rate of 30 Hz giving an indicator for a real-time capable algorithm. 

We intentionally do not consider datasets containing images captured by moving cameras (e.g. MOT17). This would require an additional time-consuming motion compensation that is not necessary in our targeted scenario.

\subsection{Related Work}
\label{sec:rel}

\def\rtcol{green!80!black}
\def\slowcol{blue}
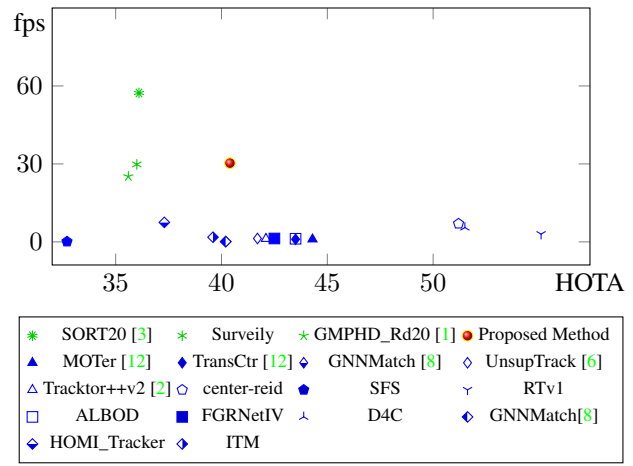
\begin{figure}
\centering
\begin{tikzpicture}
	\begin{axis}[
		x label style={at={(axis description cs:1,0.16)},anchor=north},
		y label style={at={(axis description cs:0.13,0.85)},rotate=-90,anchor=south},
		width=0.5\textwidth,
		height=5cm,
		ymax=90,xmin=32,
		legend style={
			at={(0.5,-0.2)},
			anchor=north,
			legend columns=4
		},
		ytick={0,30,60},
		xtick={35, 40, 45, 50},
		xlabel = {HOTA},
		ylabel = {fps}],
		\addplot+[
			nodes near coords,only marks,
			scatter/classes={
				SORT20={mark=10-pointed star,\rtcol},
				Surveily={mark=asterisk, \rtcol},
				GMPHD_Rd20={mark=star, \rtcol},
				Our={mark=ball, yellow, ball color=red},
				MOTer={mark=triangle*, \slowcol},
				TransCtr={mark=diamond*, \slowcol},
				GNNMatch={mark=halfdiamond*, \slowcol},
				UnsupTrack={mark=diamond, \slowcol},
				Tracktor++v2={mark=triangle, \slowcol},
				center-reid={mark=pentagon, \slowcol},
				SFS={mark=pentagon*, \slowcol},
				RTv1={mark=Mercedes star flipped, \slowcol},
				ALBOD={mark=square, \slowcol},
				FGRNetIV={mark=square*, \slowcol},
				D4C={mark=Mercedes star, \slowcol},
				GNNMatch={mark=halfsquare left*, \slowcol},
				HOMI_Tracker={mark=halfsquare*, \slowcol},
				ITM={mark=halfsquare right*, \slowcol}
			},
			point meta=explicit symbolic
			]
		table[meta=name] {
			x y label name
			36.1 57.3 \scriptsize\cite{bewley2016simple} SORT20
			36.0 29.8 \scriptsize S Surveily
			35.6 25.2 \scriptsize\cite{baisa2020occlusionrobust} GMPHD_Rd20
			40.4 30.3 \scriptsize Our Our
			44.3 1.0 \scriptsize\cite{xu2021transcenter} MOTer
			43.5 1.0 \scriptsize\cite{xu2021transcenter} TransCtr
			40.2 0.1 \scriptsize\cite{papakis2021gcnnmatch} GNNMatch
			41.7 1.3 \scriptsize\cite{karthik2020simple} UnsupTrack
			42.1 1.2 \scriptsize\cite{bergmann2019tracking} Tracktor++v2
			51.2 7.0 \scriptsize center-reid center-reid
			32.7 0.1 \scriptsize SFS SFS
			55.1 3.0 \scriptsize RTv1 RTv1
			43.5 1.2 \scriptsize ALBOD ALBOD
			42.5 1.3 \scriptsize FGRNetIV FGRNetIV
			51.5 5.5 \scriptsize D4C D4C
			40.2 0.1 \scriptsize\cite{papakis2021gcnnmatch} GNNMatch
			37.3 7.5 \scriptsize HOMI_Tracker HOMI_Tracker
			39.6 1.8 \scriptsize ITM ITM
		};
	\legend{\scriptsize SORT20~\cite{bewley2016simple},
	 		\scriptsize Surveily,
 			\scriptsize GMPHD\_Rd20~\cite{baisa2020occlusionrobust},
 			\scriptsize Proposed Method,
 			\scriptsize MOTer~\cite{xu2021transcenter},
 			\scriptsize TransCtr~\cite{xu2021transcenter},
 			\scriptsize GNNMatch~\cite{papakis2021gcnnmatch},
 			\scriptsize UnsupTrack~\cite{karthik2020simple},
 			\scriptsize Tracktor++v2~\cite{bergmann2019tracking},
 			\scriptsize center-reid,
 			\scriptsize SFS,
 			\scriptsize RTv1,
 			\scriptsize ALBOD,
 			\scriptsize FGRNetIV,
 			\scriptsize D4C,
 			\scriptsize GNNMatch\cite{papakis2021gcnnmatch},
 			\scriptsize HOMI\_Tracker,
 			\scriptsize ITM
 			}
	\end{axis}
\end{tikzpicture}
\caption{Approaches solving the MOT20 benchmark with focus on runtime vs. Higher Order Tracking Accuracy (HOTA)~\cite{Luiten_2020}. A clear gap can be seen between real-time (green) and non-real-time approaches (blue). The red dot indicates the proposed approach.}
\label{fig:hota_cmp}
\end{figure}

Evaluations of over 20 different approaches are available on MOT20. As depicted in Fig.~\ref{fig:hota_cmp}, a significant gap divides two clusters of algorithms regarding the runtime given by the authors.
These algorithms are evaluated on different systems and thus the definition of real-time can only be vague. 
Furthermore, the execution time also depends on the number of detections. 
Thus, within our development we focus on linear runtime with respect to the big O notation.
For comparison with other approaches based on MOT20, we define real-time capability based on the gap in Fig.~\ref{fig:hota_cmp}.
One cluster can be seen below the gap with varying performance regarding High Order Tracking Accuracy (HOTA).
We define the algorithms belonging to the other cluster above the gap as realtime capable, 
although not all are above 30 fps which is the frame rate of MOT20. 
Solutions belonging to this cluster rely on the detections given in MOT20.
To remain fast, one cannot expand those algorithms by complex image processing. 
Faster RCNN~\cite{conf/nips/RenHGS15} is used to provide detections in MOT20, but it reveals a weak performance in crowded test images. 
Thus, the performance of real-time capable approaches is rather low.

This is even more obvious when the solutions are sorted by the MOTA metric. 
All real-time capable solutions perform below all non-real-time approaches, see Table~\ref{tab:res}.

SORT~\cite{bewley2016simple} is an example of a simple but fast approach. It applies a Kalman filter for tracking that is updated with detection bounding boxes. The assignment is done by applying the Intersection over Union (IoU) distance to build a cost matrix solved by the Hungarian algorithm. However, as this approach ignores appearance features, it is fast but tracking performance is rather low (see Fig.~\ref{fig:hota_cmp}).

Baisa~\cite{baisa2020occlusionrobust} proposes to improve tracking performance by applying an identification network (IdNet) that extracts features from detections. A GM-PHD filter first uses detections to output estimates which are then used for an estimate to track association. Two disadvantages are worth mentioning here: 1) Different and inconsistent distance metrics are applied throughout the pipeline and 2) IdNet is trained on single images instead of the (dis)similarity of two patches.

Using a CNN for similarity estimation is a common approach. Ding et al.~\cite{ding2015deep} propose to build triplets for training a CNN that extracts feature representations from image patches. Siamese networks are widely used in single object tracking~\cite{pflugfelder2017depth} and person re-identification~\cite{varior2016gated}.

\subsection{Approach} 

The base of the proposed tracker is similar to SORT. I.e. we apply Kalman filter, one for each track, and update them utilizing detections. For our targeted scenario, this solution is sufficiently fast but lacks accuracy due to erroneous detections.
We thus improve this approach by applying an additional feature extraction from image method. Siamese networks could help to improve the association of possibly erroneous detections to tracks.
However, Siamese networks applied for tracking usually have a $O(N\cdot M) \approx O(N^2)$ runtime, where $N$ is the number of tracks and $M$ the number of detections. This is especially problematic in a 24/7 surveillance scenario.


Our contribution is LTSiam, a CNN that
\begin{itemize}
	\item is based on well-evaluated and well-performing Siamese networks,
	\item is trained with the same similarity measure used for inference,
	\item is specifically trained for multi object tracking application,
	\item can be partially applied for linear instead of square complexity and
	\item can be applied in an online and real-time capable algorithm.
\end{itemize}

In the evaluation, we can show that this approach outperforms other real-time approaches in HOTA, MOTA and IDF1 on the MOT20 dataset while maintaining real-time performance.

\section{Approach}

In this paper, we assume that detections are given from an external source like a CNN detector. We thus exclude this step from our timing analysis as a second system could be utilized for obtaining detections in parallel. 

\subsection{Track to Detection Assignment}

For each person tracked we apply a Kalman filter. This allows us to continue tracking even if a person is not detected for some frames. Thus, detections must be associated with a Kalman filters. We do this by creating an $N \times M$ cost matrix $C$ where a single value $c_{n,m}$ expresses a cost for assigning detection $m$ to track $n$. Afterwards, we utilize the Hungarian algorithm to minimize the overall cost and to output a set of selected associations $A=\{\left(m_1, n_1\right),\dots\}$.

This is a multi-criteria optimization consisting of the Intersection over Union $c^{IoU}$, normalized distance $c^d$ and the cost $c^{f}$ of the fingerprint similarity (see next section):

\begin{equation}
	\label{eq:cost_entries}
	c_{n,m} = c^{IoU}_{n,m} + \alpha \cdot c^{d}_{n,m} + \beta \cdot c^{f}_{n,m},
\end{equation}

where $\alpha$ and $\beta$ are weights heuristically determined.

The normalized distance $c^d$ is given by the pixel distance of track $n$ to detection $m$ divided by the maximum possible distance in the image, which is the distance of opposite image corners. 

\subsubsection{Appearance of new untracked persons}

Let us assume a person enters the observed area with detection $j$. Two cases can occur: 1) The detection is not assigned to any existing track which can and should happen if $N < M$, 2) detection $j$ is assigned to an existing track $i$. The second case can occur if another person $i$ left the observed area at the same time. To handle this, if

\begin{equation}
	\label{eq:thres}
	c_{i,j} > \Lambda_c
\end{equation}

we assume that this assignment is wrong, where $\Lambda_c$ is a heuristically determined threshold. In this case the assignment $\left(i, j\right)$ is removed from $A$. Afterwards, for all detections not in $A$ new tracks are created.

\subsubsection{Disappearance of tracked persons}
\label{sec:dis}

In case a track is not contained in $A$ (i.e. no detection is assigned to this track in this frame) there are three possible reasons: 1) the person finally left the observed area, 2) it is temporarily hidden and 3) it is a false negative detection. Cases 2) and 3) cannot be distinguished and are thus handled equally by continuing the track (without sensor updates). To handle case 1) the track is deleted if it did not get any updates for $T$ frames.

\subsection{Structure of LTSiam}
\newcommand{\framedpic}[1]{%
	\setlength{\fboxsep}{0pt}%
	\setlength{\fboxrule}{1pt}%
	\fbox{\includegraphics[width=.03\textwidth]{#1}}}%
\begin{figure}
\centering
\begin{tikzpicture}

	\node[inner sep=0pt] (PA1) at (1.0,0.2)
	{\framedpic{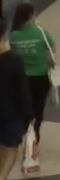}};
	\node[inner sep=0pt] (PA2) at (0.5,0.1)
	{\framedpic{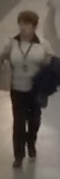}};
	\node[inner sep=0pt] (PA3) at (0,0)
	{\framedpic{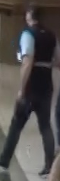}};
	\node[draw,minimum width=2cm, minimum height=1cm, below=0.5cm of PA2] (VGG) {VGG16};
	\node[draw,minimum width=1.8cm, below=0cm of VGG] (FC1) {FC 4096};
	\node[draw,minimum width=1.8cm, below=0cm of FC1] (RELU) {ReLU};
	\node[draw,below=0cm of RELU] (FC2) {FC 100};
	\draw[->,thick] (PA2.south) -- (VGG.north) node[pos=0.5,right] {$60 \times 35 \times 3$};
	\node[draw] (COS) at (3.0,-4.5) {Cosine Similarity};
	\draw[->,thick] (FC2.south) |- (COS.west) node[midway,fill=white] {$F_A$};
	
	\node[inner sep=0pt] (PB1) at (6,0.2)
	{\framedpic{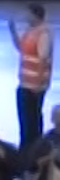}};
	\node[inner sep=0pt] (PB2) at (5.5,0.1)
	{\framedpic{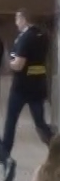}};
	\node[inner sep=0pt] (PB3) at (5.0,0)
	{\framedpic{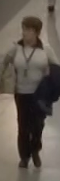}};
	\node[draw,minimum width=2cm, minimum height=1cm, below=0.5cm of PB2] (VGG) {VGG16};
	\node[draw,minimum width=1.8cm, below=0cm of VGG] (FC1) {FC 4096};
	\node[draw,minimum width=1.8cm, below=0cm of FC1] (RELU) {ReLU};
	\node[draw,below=0cm of RELU] (FC2) {FC 100};
	\draw[->,thick] (PB2.south) -- (VGG.north);
	\draw[->,thick] (FC2.south) |- (COS.east) node[midway,fill=white] {$F_B$};
	\node[draw,below=0.3cm of COS] (SQR) {$y=x^2$};
	\coordinate[below=1cm of SQR] (output);
	\draw[->,thick] (COS.south) -- (SQR.north);
	\draw[->,thick] (SQR.south) -- (output.north) node[midway,fill=white] {$y \in \left[0, \dots, 1\right]$};
	\draw[very thick,fill=black] (2.75,-2) rectangle (3.25,-2.1);
	\draw[very thick,fill=black] (2.75,-2.2) rectangle (3.25,-2.3);
\end{tikzpicture}
\caption{Complete LTSiam model used for training. Input for a VGG backbone is a small image patch. A first fully connected layer gives a feature vector consisting of 4096 values. Another fully connected layer shortens this to 100 values to ensure a short runtime of the cosine similarity. The latter has a complexity of $O(n^2)$ during inference. After squaring the output 0 means "dissimilar" and 1 "similar".}
\label{fig:ltsiam}
\end{figure}
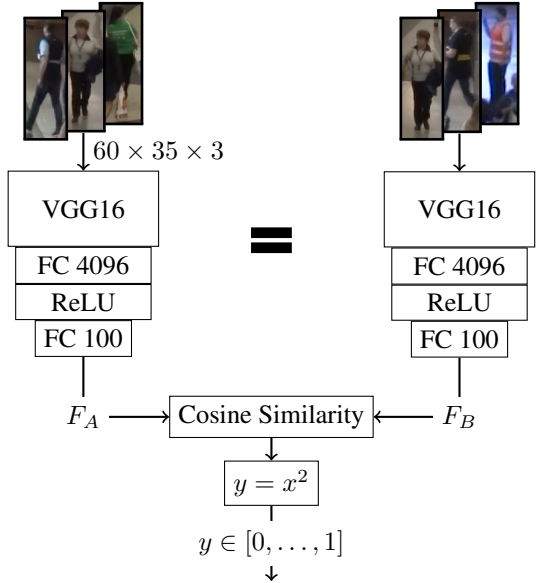

Fig.~\ref{fig:ltsiam} depicts the proposed LTSiam network. 
%
%
As the setup of the cost matrix (see Eq.~\ref{eq:cost_entries}) has necessarily a quadratic complexity we limit the required calculations for this to a minimum. Therefore, the comparison of the fingerprints $F_A$ and $F_B$ is realized by a simple cosine similarity. The result is -1 for diametrically opposed vectors, 0 for vectors oriented at $90^{\circ}$ relative to each other, and 1 for same orientation. However, interpreting -1 as dissimilar and 1 as similar patches (with 0 in between) does not lead to adequate training results. Thus, the similarity is squared, so both diametrically opposed and same oriented vectors are interpreted as similar patches. Given this network setup, the training leads to satisfying results and additionally opens up the possibility for inference with linear complexity.

To achieve this, we split off the backbone including fully connected layers. 
This can be utilized to infer the fingerprint at complexity $O(M)$. 
The result is saved in the track to which the detection was assigned and can be reused in the next frame.
The only remaining part with squared complexity is then the application of the fingerprints for determining the squared cosine similarity. 

Note that the fingerprints are inferred for image patches derived from detections only. These patches thus do not depend on the tracking results. This is an important property, because GPUs are fast in processing large batches of images, but to run the inference an overhead in the calculation time compromises the real-time capability. We thus buffer detections for 1-2 s and run the inference then once. This hides the overhead due to initialization sufficiently. Although the tracking results are then delayed about this buffer length, it is still an online algorithm as results are continuously provided during runtime.

For training the network, we utilize the training scenes of the MOT20 and MOT17\footnote{Note that moving cameras are only problematic for the Kalman filter. Training image patch similarities are not affected by this.} datasets providing 3856 annotated tracks. From this, we extract 1437801 patches from detections with resolution $35 \times 60$.  
Each training batch consists of 50\% pairs showing the same person and 50\% showing different persons. 
We only use pairs from the same scene as otherwise the background from different scenes would obviously indicate different persons.
Furthermore, in contrast to Siamese networks for reidentification, the temporal distance between image pairs is at most the timeout $T$, see section~\ref{sec:dis}. Thus, we limit the temporal distance during training to 50 frames for a pair\footnote{We do not limit it to timeout $T$ as this value may change after training.}. Due to the large number of possible pairs under these constraints (up to $10^{12}$) we generate pairs randomly during training.

Training is performed with a batch size of 50 in 1000000 steps. 
The mean average error is minimized utilizing stochastic gradient descent.
\section{Evaluation}
\label{sec:eval}
\catcode`\_=12
\begin{table}
\sisetup{input-open-uncertainty= ,  
		input-close-uncertainty= , 
		table-space-text-pre=(,    
		table-space-text-post=),   
		table-align-text-pre=false,
		detect-weight,
		mode=text,                 
		table-format=-1.2}
\renewcommand{\bfseries}{\fontseries{b}\selectfont} 
\newrobustcmd{\B}{\bfseries}  
\begin{center}
\setlength\tabcolsep{2pt}
\begin{tabular}{|l|c|c|c|c|c|}%
	\hline
	\bfseries Short & \bfseries HOTA & \bfseries MOTA & \bfseries IDF1 & \bfseries MOTP & \bfseries RT (s)
	\begin{filecontents*}{MOT20_results.csv}
short_name,sequence,MOTA,IDF1,HOTA,MOTP,MT,ML,FP,FN,Rcll,Prcn,AssA,DetA,AssRe,AssPr,DetRe,DetPr,LocA,FAR,IDSW,FM,Rcll,Prcn,runtime,IDP,IDR,IDSWR,detector
XJTU_priv,OVERALL,64.2994,66.6209,50.5445,74.7589,626,174,40780,140565,72.8338,90.2356,50.8206,50.8656,54.6935,74.3986,56.3587,69.8242,78.4632,9.10471,3379,7405,72.8338,90.2356,135.73,74.5796,60.197,,1
XJTU_priv,MOT20-08,36.2475,47.7036,37.1284,72.8358,46,51,14801,34056,56.0477,74.5814,38.9484,35.8877,42.6213,66.6758,43.1351,57.3989,76.3981,18.3635,541,763,56.0477,74.5814,135.73,55.5909,41.7764,,1
XJTU_priv,MOT20-07,75.6563,57.2538,50.9667,81.5673,83,2,2564,4924,85.1243,91.6593,41.2219,63.8907,45.3905,72.618,71.3721,76.8514,83.7114,4.38291,570,723,85.1243,91.6593,135.73,59.4515,55.2128,,1
XJTU_priv,MOT20-06,48.5571,53.1018,40.831,75.5653,77,87,8096,59011,55.5496,90.1078,41.7004,40.2812,45.0423,71.2201,43.6176,70.7528,78.8836,8.03175,1187,2663,55.5496,90.1078,135.73,69.6195,42.919,,1
XJTU_priv,MOT20-04,78.4832,78.2668,57.5265,74.0341,420,34,15319,42574,84.4668,93.7937,56.3324,59.1473,60.1639,75.5743,64.4552,71.5724,77.9916,7.3649,1081,3256,84.4668,93.7937,135.73,82.5879,74.3754,,1
UnsupTrack,OVERALL,53.633,50.5592,41.6734,80.0902,376,311,6439,231298,55.2983,97.7991,40.2369,43.3185,43.2107,75.9211,45.1151,79.7958,82.5709,1.4376,2178,4335,55.2983,97.7991,3467.3,69.9884,39.5734,39.3864,0
UnsupTrack,MOT20-08,23.9598,28.988,24.1185,77.2002,11,106,1936,56628,26.9165,91.5058,28.5431,20.5608,30.7573,71.7802,21.3683,72.6535,80.1577,2.40199,355,780,26.9165,91.5058,3467.3,63.768,18.7574,13.1889,0
UnsupTrack,MOT20-07,49.781,47.1298,38.2191,80.3514,23,25,140,16310,50.7266,99.1731,36.904,39.9854,39.6898,75.9085,41.4118,80.9624,83.0336,0.239316,173,357,50.7266,99.1731,3467.3,69.6356,35.6183,3.41044,0
UnsupTrack,MOT20-06,31.4665,32.0349,25.8478,77.964,24,124,1796,88455,33.3707,96.104,26.1716,25.6617,27.8229,71.9207,26.5191,76.3935,80.8721,1.78175,732,1546,33.3707,96.104,3467.3,62.1459,21.5793,21.9354,0
UnsupTrack,MOT20-04,73.2235,62.3052,51.0576,80.8253,318,56,2567,69905,74.495,98.7584,44.5775,58.6309,47.8783,77.0599,61.2828,81.246,83.1424,1.23413,918,1652,74.495,98.7584,3467.3,72.4517,54.6515,12.323,0
UniMOT,OVERALL,70.3198,69.1075,54.244,78.0167,700,155,25361,125631,75.72,93.9205,52.8414,55.938,57.6819,72.9428,60.6133,75.231,80.9934,5.6622,2581,6657,75.72,93.9205,749.08,77.413,62.4114,34.0861,1
UniMOT,MOT20-08,52.5296,53.6338,42.2634,77.3137,51,48,4751,31397,59.4794,90.6546,41.7821,42.9673,47.8841,64.4549,46.9994,71.7449,80.4854,5.89454,634,1422,59.4794,90.6546,749.08,67.6895,44.4118,10.6592,1
UniMOT,MOT20-07,77.8405,70.4545,56.348,78.8188,84,2,2056,5074,84.6712,93.1656,51.2972,62.5797,57.2995,70.638,68.8764,75.7888,81.6893,3.51453,205,243,84.6712,93.1656,749.08,73.9886,67.2427,2.42113,1
UniMOT,MOT20-06,56.1243,52.3133,43.0378,78.4607,105,79,7280,49891,62.4193,91.9242,40.0504,46.4705,45.7385,62.7371,50.3974,74.3096,81.2366,7.22222,1077,2192,62.4193,91.9242,749.08,64.6773,43.9178,17.2543,1
UniMOT,MOT20-04,81.3167,79.9576,61.393,77.9023,460,26,11274,39269,85.6726,95.4187,59.5995,63.4838,63.7213,78.3097,68.4122,76.2193,80.9126,5.42019,665,2800,85.6726,95.4187,749.08,84.5056,75.8742,7.7621,1
TTS,OVERALL,77.4646,75.2363,61.9198,81.7381,878,113,34201,80788,84.3866,92.7362,60.1279,63.9555,66.3021,74.5158,70.64,77.6295,83.9123,7.63586,1615,2421,84.3866,92.7362,1740,78.9584,71.8493,,1
TTS,MOT20-08,61.0255,60.8081,48.6414,79.3773,83,31,7814,21939,71.6858,87.6671,46.5058,51.1432,53.489,64.5123,58.205,71.181,81.9784,9.69479,446,714,71.6858,87.6671,1740,67.5863,55.2656,,1
TTS,MOT20-07,77.8134,72.9398,59.041,80.8294,84,6,2711,4490,86.4355,91.3447,55.3519,63.3041,61.2969,72.1804,71.4226,75.4791,83.298,4.63419,143,211,86.4355,91.3447,1740,75.0112,70.9797,,1
TTS,MOT20-06,64.3446,61.8916,50.8586,80.3011,139,52,10288,36309,72.65,90.3613,48.0454,54.0136,53.5966,67.9414,59.987,74.6112,82.6249,10.2063,738,956,72.65,90.3613,1740,69.4358,55.8261,,1
TTS,MOT20-04,88.4247,85.1277,70.2347,82.8931,572,24,13388,18050,93.4144,95.0308,67.856,72.8242,74.0548,79.1271,79.2209,80.5917,84.8784,6.43654,288,540,93.4144,95.0308,1740,85.8642,84.4037,,1
TrTEST,OVERALL,56.9935,48.7944,42.4692,79.7144,499,243,26770,190486,63.1858,92.4317,37.9845,47.8909,40.4246,79.0078,51.8072,75.8205,82.3414,5.97678,5271,4877,63.1858,92.4317,4526,60.0868,41.0751,83.4206,0
TrTEST,MOT20-08,26.3913,30.4246,25.0707,74.971,21,84,8087,48069,37.9627,78.4358,23.8749,26.4872,25.2198,73.3135,29.6677,61.3725,78.5265,10.0335,879,890,37.9627,78.4358,4526,46.6428,22.575,23.1543,0
TrTEST,MOT20-07,55.5874,48.9348,41.0533,80.9919,38,16,598,13615,58.8683,97.0225,36.5497,46.5268,38.632,80.3504,48.823,80.4665,83.5386,1.02222,488,402,58.8683,97.0225,4526,64.7929,39.313,8.28969,0
TrTEST,MOT20-06,32.1625,28.9787,24.3349,74.0675,37,100,11487,76661,42.2546,83.0031,20.0546,29.8809,21.2332,73.0784,32.8927,64.6866,77.8341,11.3958,1911,1887,42.2546,83.0031,4526,42.9516,21.8655,45.2258,0
TrTEST,MOT20-04,77.8418,60.8804,52.7374,81.6581,403,43,6598,52141,80.9763,97.113,43.8269,63.7331,46.5994,80.4335,67.5881,81.0696,83.8622,3.17212,1993,1698,80.9763,97.113,4526,66.9464,55.8223,24.6121,0
TransCtr,OVERALL,61.0433,49.8437,43.5031,79.5358,601,192,49189,147890,71.4181,88.2527,36.1002,52.9159,44.533,58.9852,59.1549,73.0988,81.7028,10.9821,4493,8950,71.4181,88.2527,4478.5,55.7182,45.0897,,0
TransCtr,MOT20-08,31.4581,31.5708,26.4886,76.7515,35,64,13050,39038,49.618,74.6582,21.1886,33.7404,26.8158,46.0444,39.95,60.1112,79.3409,16.1911,1021,1941,49.618,74.6582,4478.5,39.5371,26.2764,,0
TransCtr,MOT20-07,76.5415,58.7818,50.8837,81.3623,71,5,1372,6106,81.5534,95.1634,42.0696,62.1181,49.0502,62.725,67.4376,78.6919,83.3051,2.3453,287,448,81.5534,95.1634,4478.5,63.6867,54.5784,,0
TransCtr,MOT20-06,45.581,35.6408,30.7229,78.038,89,72,14441,55970,57.8403,84.1704,23.0217,41.4552,29.2917,45.5223,46.9006,68.2508,80.6311,14.3264,1834,3210,57.8403,84.1704,4478.5,43.753,30.0662,,0
TransCtr,MOT20-04,75.0248,59.4057,51.4755,80.2958,406,51,20326,46776,82.9337,91.7919,42.0271,63.4724,51.7166,64.918,69.5195,76.9449,82.2674,9.77212,1351,3351,82.9337,91.7919,4478.5,62.5782,56.5392,,0
TransCenter,OVERALL,61.9472,50.415,44.3284,79.8738,614,193,45895,146347,71.7163,88.9933,37.1139,53.4151,45.221,59.7739,59.49,73.8216,82.0311,10.2467,4653,9462,71.7163,88.9933,4478.5,56.4877,45.5213,,1
TransCenter,MOT20-08,35.0679,30.8228,26.861,77.0612,36,67,9550,39617,48.8707,79.8595,21.5576,34.1894,25.952,46.4226,39.1389,63.9569,79.8954,11.8486,1145,2312,48.8707,79.8595,4478.5,40.5951,24.8425,,1
TransCenter,MOT20-07,75.4358,61.4855,51.8168,81.5366,80,7,2173,5687,82.8193,92.6556,43.2034,62.4649,51.6488,61.5903,69.1253,77.3352,83.6718,3.71453,271,417,82.8193,92.6556,4478.5,65.1367,58.2218,,1
TransCenter,MOT20-06,46.0194,35.3425,31.346,78.701,88,75,12947,56915,57.1284,85.4182,24.0315,41.4384,30.0244,47.3104,46.5151,69.5492,81.2229,12.8442,1801,3443,57.1284,85.4182,4478.5,44.0933,29.49,,1
TransCenter,MOT20-04,75.6319,60.1098,52.2603,80.5256,410,44,21225,44128,83.8998,91.5499,43.0061,63.9411,52.2744,65.5993,70.3643,76.7802,82.4443,10.2043,1436,3290,83.8998,91.5499,4478.5,62.8503,57.5984,,1
Tracktor++v2,OVERALL,52.6004,52.6726,42.1009,79.8679,365,331,6930,236680,54.2582,97.591,42.048,42.2767,45.8839,71.6292,44.0942,79.313,82.3765,1.54722,1648,4374,54.2582,97.591,3795,73.7058,40.9786,30.3733,0
Tracktor++v2,MOT20-08,21.0043,27.1723,23.4273,78.8388,12,116,2078,58880,24.0101,89.9526,29.4984,18.723,32.0304,66.8862,19.449,72.868,81.0918,2.57816,251,734,24.0101,89.9526,3795,64.486,17.2126,10.4539,0
Tracktor++v2,MOT20-07,50.077,49.5877,39.2749,81.0773,27,23,252,16127,51.2794,98.5371,38.4359,40.3928,42.3625,72.4187,42.0552,80.812,83.4751,0.430769,146,304,51.2794,98.5371,3795,72.437,37.6967,2.84715,0
Tracktor++v2,MOT20-06,30.1235,33.2424,26.5512,78.7918,28,137,1745,90509,31.8236,96.0335,28.6054,24.7083,31.3539,65.8971,25.5131,76.9991,81.4116,1.73115,512,1408,31.8236,96.0335,3795,66.7788,22.1292,16.0887,0
Tracktor++v2,MOT20-04,72.7244,65.3696,51.4903,80.0852,298,55,2855,71164,74.0357,98.6126,46.21,57.531,50.35,73.0939,60.3078,80.3299,82.6048,1.3726,739,1928,74.0357,98.6126,3795,76.2197,57.2237,9.98167,0
TMM,OVERALL,43.3109,45.207,36.1733,78.201,218,327,27953,262406,49.2863,90.1217,35.7713,36.979,39.5758,65.2076,39.8231,72.8179,80.7934,6.2409,2965,7477,49.2863,90.1217,20338200,63.9347,34.965,,1
TMM,MOT20-08,13.3486,24.3303,20.0594,70.4661,10,101,11155,55438,28.4523,66.4016,20.928,19.6169,23.4757,51.561,22.2496,51.9257,74.1034,13.84,548,1106,28.4523,66.4016,20338200,40.556,17.3778,,1
TMM,MOT20-07,48.4457,47.2673,36.7295,77.4947,31,22,1216,15619,52.8141,93.4966,35.0763,39.1567,39.0948,66.6068,41.848,74.0834,80.6967,2.07863,230,398,52.8141,93.4966,20338200,65.4722,36.9838,,1
TMM,MOT20-06,24.2089,29.9431,24.2228,72.9109,24,123,12359,87171,34.3379,78.6712,24.4835,24.3737,27.2384,55.7937,26.7377,61.2585,76.2012,12.2609,1088,1955,34.3379,78.6712,20338200,49.2726,21.5062,,1
TMM,MOT20-04,60.4136,56.6441,44.1538,80.6966,153,81,3223,104178,61.9905,98.1384,40.2054,48.7095,44.3,68.45,50.8847,80.5566,82.9901,1.54952,1099,4018,61.9905,98.1384,20338200,73.1593,46.2121,,1
Surveily,OVERALL,44.6093,42.5106,36.0229,76.1165,393,236,71208,211064,59.2089,81.1404,32.0259,41.2117,36.2729,62.3749,47.4669,65.0817,78.797,15.8982,4334,6646,59.2089,81.1404,150.5,50.3838,36.7654,73.1985,0
Surveily,MOT20-08,5.43209,24.5409,20.6973,69.1464,13,81,21850,50637,34.6484,55.1307,20.0739,21.9536,23.2521,47.9825,27.4673,43.7575,72.6485,27.1092,788,1237,34.6484,55.1307,150.5,31.7946,19.9822,22.7427,0
Surveily,MOT20-07,47.7418,47.2116,37.127,75.4806,36,15,3824,13184,60.1704,83.8928,33.772,41.571,38.5952,63.9295,47.3519,66.0512,78.7607,6.53675,290,403,60.1704,83.8928,150.5,56.5183,40.5365,4.81965,0
Surveily,MOT20-06,19.0709,27.8928,23.5766,70.9443,34,90,28598,77325,41.7545,65.9669,20.6075,27.6577,23.9853,50.1155,33.0515,52.254,74.2484,28.371,1516,2278,41.7545,65.9669,150.5,35.98,22.7739,36.3075,0
Surveily,MOT20-04,67.6763,52.9494,44.3181,78.4993,310,50,16936,69918,74.4903,92.3402,36.0781,54.8998,40.5127,66.5512,60.1169,74.5428,80.9896,8.14231,1740,2728,74.4903,92.3402,150.5,59.2934,47.8317,23.3587,0
SORT20,OVERALL,42.6614,45.1287,36.1034,78.4897,208,326,27521,264694,48.8441,90.1799,35.8846,36.7068,39.4308,66.9142,39.5162,72.9606,80.9847,6.14445,4470,17798,48.8441,90.1799,78.2,64.2245,34.7858,91.5157,0
SORT20,MOT20-08,13.0659,24.2229,19.8607,70.8842,13,102,10974,55559,28.2962,66.6434,20.6276,19.5032,23.0787,52.6718,22.1121,52.0834,74.3386,13.6154,827,2091,28.2962,66.6434,78.2,40.6365,17.2539,29.2266,0
SORT20,MOT20-07,48.4668,47.2735,36.7023,77.6406,30,22,1032,15666,52.6721,94.4117,35.1002,39.0515,38.878,67.529,41.5993,74.5724,80.7957,1.7641,360,872,52.6721,94.4117,78.2,66.0042,36.8237,6.83473,0
SORT20,MOT20-06,23.6944,29.5447,23.8759,73.071,26,123,12309,87352,34.2016,78.6724,23.9813,24.1464,26.5855,56.5094,26.526,61.0209,76.3012,12.2113,1640,3982,34.2016,78.6724,78.2,48.7525,21.1944,47.951,0
SORT20,MOT20-04,59.5139,56.7295,44.1809,81.0354,139,79,3206,106117,61.283,98.127,40.5317,48.3617,44.315,70.5603,50.4768,80.8244,83.2134,1.54135,1643,10853,61.283,98.127,78.2,73.7827,46.0793,26.81,0
SFS,OVERALL,50.7999,41.147,32.7101,74.8796,341,251,50139,200932,61.167,86.3245,25.072,43.0517,34.5384,47.4389,47.974,67.7905,78.2079,11.1942,3503,7617,61.167,86.3245,44500,49.6087,35.1513,57.2694,0
SFS,MOT20-08,23.8707,31.9632,24.5617,69.3347,21,70,14008,44270,42.8656,70.3359,21.8713,28.0476,25.7478,53.0925,32.734,53.8997,73.7533,17.3797,710,1282,42.8656,70.3359,44500,42.2049,25.7214,16.5634,0
SFS,MOT20-07,59.394,54.4069,41.4153,72.4768,43,9,2695,10511,68.2457,89.3415,37.1156,46.7075,42.0413,62.3151,51.495,67.4663,76.4608,4.60684,235,501,68.2457,89.3415,44500,62.8159,47.9834,3.44344,0
SFS,MOT20-06,33.7059,31.7806,26.3658,71.9241,46,97,18710,68212,48.6189,77.5269,21.0535,33.5152,27.4496,43.9436,37.9987,60.7191,75.4746,18.5615,1088,1945,48.6189,77.5269,44500,41.2288,25.8555,22.3781,0
SFS,MOT20-04,65.6547,46.0843,36.3308,77.0678,231,75,14726,77939,71.5638,93.0166,25.4071,52.2348,37.195,45.7285,56.6888,73.7181,80.1464,7.07981,1470,3889,71.5638,93.0166,44500,52.9916,40.77,20.5411,0
RTv1,OVERALL,60.6369,67.9319,55.1139,78.7532,865,91,112927,85062,83.5605,79.2905,55.7667,54.7428,61.0609,73.3161,68.2973,64.8072,81.2572,25.2125,5686,7934,83.5605,79.2905,1500,66.1962,69.7611,,0
RTv1,MOT20-08,21.6187,50.6186,41.7346,76.2438,79,28,36457,23131,70.1474,59.8535,46.3893,37.9605,51.7863,67.9025,56.4267,48.1463,78.9353,45.232,1145,1838,70.1474,59.8535,1500,46.9045,54.9713,,0
RTv1,MOT20-07,74.1669,70.5705,57.5476,78.9451,84,2,3462,4753,85.6409,89.1166,54.0875,61.8892,58.8137,74.4169,70.2776,73.1298,81.7466,5.91795,336,462,85.6409,89.1166,1500,72.0025,69.1943,,0
RTv1,MOT20-06,39.1143,50.6402,42.4338,76.8297,139,43,42947,35204,73.4824,69.4327,40.6823,44.7147,45.4543,63.6123,59.6693,56.3809,79.1507,42.6062,2679,3241,73.4824,69.4327,1500,49.2448,52.117,,0
RTv1,MOT20-04,80.4582,81.3566,64.6709,80.0169,563,18,30061,21974,91.9827,89.3465,63.672,65.812,69.1004,77.897,75.593,73.4265,82.5521,14.4524,1526,2393,91.9827,89.3465,1500,80.1907,82.5568,,0
RADTrack,OVERALL,67.1501,70.4678,56.4788,79.2219,773,111,61134,104597,79.7851,87.1015,56.3677,56.8213,60.3215,77.4918,65.288,71.275,81.7594,13.649,4243,8236,79.7851,87.1015,1640,73.6988,67.5082,,1
RADTrack,MOT20-08,37.4684,54.8846,43.5289,76.7513,68,36,20273,27338,64.7179,71.2109,46.9091,40.7442,50.9267,72.7073,52.1611,57.3943,79.5785,25.1526,841,1765,64.7179,71.2109,1640,57.6379,52.3824,,1
RADTrack,MOT20-07,76.4539,72.7299,59.0686,79.3677,79,3,2286,5221,84.2271,92.4219,56.0029,62.918,59.864,77.9679,69.2379,75.9743,82.0602,3.90769,287,476,84.2271,92.4219,1640,76.268,69.5055,,1
RADTrack,MOT20-06,49.0776,55.1471,44.6042,77.6133,118,51,22616,43058,67.5663,79.8638,43.4378,46.103,46.6243,72.4343,54.795,64.768,80.0572,22.4365,1929,3233,67.5663,79.8638,1640,60.1656,50.9013,,1
RADTrack,MOT20-04,83.1712,81.5234,64.9252,80.2995,508,21,15959,28980,89.4266,93.8869,62.9289,67.1116,67.0671,80.0787,73.6045,77.2756,82.8083,7.6726,1186,2762,89.4266,93.8869,1640,83.5565,79.5869,,1
ort_track,OVERALL,69.2789,67.1521,53.9569,77.6393,765,181,31790,124468,75.9448,92.5156,52.9791,55.2671,58.0107,74.2352,60.6454,73.8779,80.7915,7.09757,2701,5370,75.9448,92.5156,233.28,74.4783,61.1382,,1
ort_track,MOT20-08,36.6501,37.4007,28.5258,69.686,22,74,3533,44717,42.2887,90.2672,28.2477,29.3755,30.9065,64.4764,31.0645,66.3087,74.8905,4.38337,836,1982,42.2887,90.2672,233.28,58.6171,27.4612,,1
ort_track,MOT20-07,80.9432,67.2297,55.8975,79.3879,84,2,1452,4661,85.9189,95.1425,48.6613,64.8014,54.0843,70.8491,70.1741,77.7075,82.1501,2.48205,195,209,85.9189,95.1425,233.28,70.8383,63.9709,,1
ort_track,MOT20-06,53.9407,49.3043,40.3949,77.5735,89,81,5426,54606,58.8677,93.5078,37.3764,43.9325,40.4812,73.1119,47.0277,74.7007,80.7314,5.38294,1115,2307,58.8677,93.5078,233.28,63.8106,40.1719,,1
ort_track,MOT20-04,84.5237,80.3409,63.5945,78.4912,570,24,21379,20484,92.5264,92.2252,61.0172,66.5131,66.7923,75.7621,74.4531,74.2108,81.4047,10.2784,555,872,92.5264,92.2252,233.28,80.2101,80.472,,1
MPNTrack,OVERALL,57.5694,59.1124,46.8432,78.9971,474,279,16953,201384,61.0797,94.9089,47.2769,46.5827,52.6649,69.9663,49.4717,76.8887,81.5628,3.785,1210,1420,61.0797,94.9089,694,75.4822,48.5774,19.8102,0
MPNTrack,MOT20-08,25.9383,36.0572,29.1805,77.2524,18,101,3757,53470,30.9922,86.4715,36.4986,23.4795,40.2792,68.0745,24.9056,69.5217,79.8327,4.66129,159,181,30.9922,86.4715,694,68.3303,24.4902,5.13032,0
MPNTrack,MOT20-07,57.4424,59.9034,46.8267,79.4861,41,17,906,13061,60.542,95.6746,48.0014,46.001,53.9623,70.2734,48.9551,77.3638,81.9845,1.54872,120,124,60.542,95.6746,694,77.2844,48.9049,1.9821,0
MPNTrack,MOT20-06,36.0448,39.7942,31.4409,77.136,39,113,4831,79649,40.0039,91.6619,32.9871,30.0897,36.5113,63.4225,31.7928,72.8651,79.9356,4.79266,425,504,40.0039,91.6619,694,65.4878,28.5808,10.624,0
MPNTrack,MOT20-04,76.9527,71.2377,56.1609,79.5953,376,48,7459,55204,79.8587,96.7045,51.71,61.1714,57.6103,71.5872,65.042,78.7783,82.1149,3.58606,506,611,79.8587,96.7045,694,78.7513,65.033,6.33619,0
MOTSOT,OVERALL,68.6087,71.3863,57.3703,79.5374,806,120,57064,101154,80.4505,87.9443,57.2972,57.67,61.6985,76.6671,65.8695,72.1176,82.0494,12.7403,4209,7568,80.4505,87.9443,20,74.711,68.3448,52.3179,1
MOTSOT,MOT20-08,43.0799,55.1159,44.5312,78.1488,68,38,16025,27216,64.8753,75.827,46.6167,42.7729,50.4132,72.7828,52.5815,61.7373,80.8639,19.8821,863,1713,64.8753,75.827,20,59.768,51.1357,13.3024,1
MOTSOT,MOT20-07,72.6353,71.1854,58.2784,80.6453,85,3,3675,5066,84.6953,88.4106,56.0355,61.069,60.5423,76.6385,70.0818,73.1861,82.7969,6.28205,317,428,84.6953,88.4106,20,72.7468,69.6897,3.74283,1
MOTSOT,MOT20-06,52.8831,55.9229,46.3303,78.485,128,52,19620,41105,69.0374,82.3675,44.7684,48.2292,48.3602,71.8387,56.3497,67.4349,80.9083,19.4643,1826,2971,69.0374,82.3675,20,61.3218,51.3977,26.4494,1
MOTSOT,MOT20-04,82.9563,82.7695,65.4145,80.0863,525,27,17744,27767,89.8692,93.2803,64.1989,66.8291,68.9694,79.1619,73.7284,76.5522,82.6429,8.53077,1203,2456,89.8692,93.2803,20,84.3404,81.2561,13.3861,1
MOTPrivate,OVERALL,57.2522,46.7259,39.6008,79.3631,444,223,42271,173903,66.3907,89.0431,32.1631,49.2243,39.9075,57.0252,54.5607,73.1767,81.7298,9.4376,5014,9285,66.3907,89.0431,4478.5,54.6973,40.7825,,1
MOTPrivate,MOT20-08,30.7018,31.0705,26.5481,76.6475,36,64,14126,38475,50.3446,73.4149,21.2305,33.876,26.4743,46.5072,40.5486,59.1299,79.1963,17.5261,1094,2101,50.3446,73.4149,4478.5,38.1895,26.1886,,1
MOTPrivate,MOT20-07,76.475,58.5558,50.5609,81.3597,71,5,1427,6074,81.6501,94.9849,41.5271,62.1039,49.0965,61.6342,67.5214,78.5488,83.2972,2.43932,286,449,81.6501,94.9849,4478.5,63.3373,54.4455,,1
MOTPrivate,MOT20-06,45.0146,34.9535,30.4291,78.0045,90,72,15432,55704,58.0406,83.314,22.5805,41.5158,28.7798,45.2742,47.198,67.75,80.5652,15.3095,1861,3304,58.0406,83.314,4478.5,42.5637,29.6519,,1
MOTPrivate,MOT20-04,68.3641,54.8946,45.3006,80.1447,247,82,11286,73650,73.1287,94.6694,36.5265,56.5203,45.2779,62.7061,60.5229,78.3505,82.4667,5.42596,1773,3431,73.1287,94.6694,4478.5,62.9794,48.6493,,1
MOTer,OVERALL,62.3026,50.2979,44.2825,79.919,606,196,43006,147505,71.4925,89.5851,37.0024,53.4527,45.2753,59.5299,59.2712,74.2709,82.092,9.6017,4545,9141,71.4925,89.5851,4478.5,56.6623,45.2188,,0
MOTer,MOT20-08,35.6164,30.6073,26.8885,77.2492,34,68,8637,40164,48.1648,81.2063,21.7523,33.9062,26.1378,46.976,38.4822,64.8815,80.0806,10.7159,1086,2149,48.1648,81.2063,4478.5,41.1058,24.3805,,0
MOTer,MOT20-07,76.0883,61.7036,51.9241,81.5375,79,7,1925,5719,82.7226,93.4316,43.1457,62.8118,51.6202,61.5076,69.0385,77.976,83.674,3.2906,271,417,82.7226,93.4316,4478.5,65.6976,58.1674,,0
MOTer,MOT20-06,46.4646,34.3833,30.7983,78.7791,88,75,12109,57202,56.9123,86.187,23.137,41.524,30.0708,45.4527,46.3561,70.201,81.318,12.0129,1761,3356,56.9123,86.187,4478.5,43.2264,28.5439,,0
MOTer,MOT20-04,75.8534,60.271,52.2938,80.5349,405,46,20335,44420,83.7933,91.866,43.0467,63.9585,52.2701,65.6864,70.2244,76.9898,82.4744,9.77644,1427,3219,83.7933,91.866,4478.5,63.1743,57.6228,,0
MOT20_TBC,OVERALL,54.4615,50.0627,0,77.3471,415,245,37937,195242,62.2667,89.4655,0,0,0,0,0,0,100,8.46997,2449,2580,62.2667,89.4655,805,60.9967,42.4528,39.3308,0
MOT20_TBC,MOT20-08,27.2237,32.9827,0,70.3978,20,73,11015,44865,42.0977,74.7559,0,0,0,0,0,0,100,13.6663,510,549,42.0977,74.7559,805,45.7762,25.7782,12.1147,0
MOT20_TBC,MOT20-07,59.0345,51.3946,0,72.8461,45,10,2225,11133,66.3666,90.8031,0,0,0,0,0,0,100,3.80342,202,223,66.3666,90.8031,805,60.8564,44.479,3.0437,0
MOT20_TBC,MOT20-06,36.4448,34.9686,0,72.4269,44,97,14844,68745,48.2174,81.1758,0,0,0,0,0,0,100,14.7262,785,810,48.2174,81.1758,805,46.9197,27.8697,16.2804,0
MOT20_TBC,MOT20-04,70.3361,60.7013,0,80.4933,306,65,9853,70499,74.2783,95.3837,0,0,0,0,0,0,100,4.73702,952,998,74.2783,95.3837,805,69.3251,53.9856,12.8167,0
MLT,OVERALL,48.8526,54.5533,43.2297,77.9803,384,274,45660,216803,58.0997,86.8143,44.1141,42.6649,46.4953,78.1762,47.2861,70.6799,80.6227,10.1942,2187,3067,58.0997,86.8143,1200,68.0342,45.5313,37.6422,1
MLT,MOT20-08,12.8969,37.0854,28.3842,70.3921,16,85,17000,50195,35.2189,61.6157,35.8421,22.6997,39.5673,64.0915,27.4207,47.9998,74.144,21.0918,296,603,35.2189,61.6157,1200,50.9833,29.1415,8.40458,1
MLT,MOT20-07,53.4939,60.5306,46.1649,76.9243,39,19,1777,13471,59.3033,91.699,49.4367,43.5127,52.9551,75.1086,47.1132,72.8872,80.1928,3.03761,146,188,59.3033,91.699,1200,77.0636,49.8384,2.46192,1
MLT,MOT20-06,27.0336,46.041,34.9061,72.3606,43,104,20121,76252,42.5627,73.7413,41.9764,29.27,45.2088,69.9025,33.5571,58.1759,75.8484,19.9613,495,887,42.5627,73.7413,1200,62.9042,36.3077,11.6299,1
MLT,MOT20-04,69.0252,62.0498,50.0716,80.7458,286,66,6762,76885,71.9484,96.6847,44.5955,56.4523,46.4854,81.4098,59.5728,80.0665,83.0192,3.25096,1250,1389,71.9484,96.6847,1200,72.7164,54.1122,17.3736,1
LPC_MOT,OVERALL,56.2558,62.527,48.9523,79.6812,424,313,11726,213056,58.8239,96.2904,52.4435,45.795,54.7453,81.256,48.0829,78.7291,82.2712,2.618,1562,1865,58.8239,96.2904,6040,82.4395,50.3624,26.5538,0
LPC_MOT,MOT20-08,25.8105,37.3503,29.3638,76.2874,13,100,3814,53380,31.1084,86.3386,37.1057,23.3779,39.0867,74.6689,24.8051,68.8839,79.2834,4.73201,291,404,31.1084,86.3386,6040,70.5065,25.404,9.3544,0
LPC_MOT,MOT20-07,50.8353,58.9032,44.4903,79.2895,20,29,229,15921,51.9018,98.6846,48.9085,40.6213,51.2423,81.0518,42.1118,80.0748,82.3321,0.391453,124,138,51.9018,98.6846,6040,85.4501,44.9412,2.38913,0
LPC_MOT,MOT20-06,35.3006,43.2276,33.3028,77.7261,35,121,3503,81891,38.3151,93.557,37.7298,29.4768,39.3815,77.9585,30.7717,75.1888,80.6349,3.4752,499,653,38.3151,93.557,6040,74.3898,30.4654,13.0236,0
LPC_MOT,MOT20-04,75.6673,75.6734,59.2228,80.567,356,63,4180,61864,77.4288,98.0684,57.8238,60.7545,60.2649,82.6419,63.7696,80.7793,83.0039,2.00962,648,670,77.4288,98.0684,6040,85.7592,67.7103,8.36898,0
LCC,OVERALL,66.0321,66.9591,53.2095,77.8027,699,165,43938,129584,74.956,89.824,52.945,53.756,56.9296,76.0441,60.0315,72.0033,80.7436,9.80978,2237,4154,74.956,89.824,20030,73.6,61.4175,29.8442,1
LCC,MOT20-08,41.9403,51.2027,38.8005,76.4134,40,54,7670,36918,52.354,84.099,40.625,37.2359,43.3043,73.2453,41.617,67.0127,79.4128,9.51613,399,842,52.354,84.099,20030,66.7261,41.5389,7.62119,1
LCC,MOT20-07,74.0521,65.0802,53.4051,78.0174,78,4,3048,5292,84.0126,90.1222,48.8321,59.2525,54.0555,71.4304,67.1571,72.0527,80.8256,5.21026,249,309,84.0126,90.1222,20030,67.4466,62.8742,2.96384,1
LCC,MOT20-06,47.0122,49.0841,38.9348,75.7339,81,77,10858,58664,55.811,87.2185,38.0893,40.0432,41.2341,73.0881,44.0751,68.9976,79.0306,10.7718,823,1550,55.811,87.2185,20030,62.8951,40.2465,14.7462,1
LCC,MOT20-04,81.0868,78.0193,61.8787,78.6327,500,30,22362,28710,89.5251,91.6477,59.7302,64.3347,63.954,77.7657,72.1055,73.8473,81.4809,10.751,766,1453,89.5251,91.6477,20030,78.9442,77.1158,8.55626,1
LBT,OVERALL,59.4767,30.2392,30.357,78.2612,441,221,18135,181364,64.9488,94.88,19.0687,49.2343,19.5889,82.8538,52.3925,76.5372,81.3041,4.04889,10179,9991,64.9488,94.88,601,37.207,25.4695,,1
LBT,MOT20-08,31.8195,15.0854,14.0958,74.2976,14,84,4010,47108,39.2029,88.3383,7.07967,28.6529,7.17439,79.1121,30.5463,68.8317,78.1681,4.97519,1711,1692,39.2029,88.3383,601,24.5391,10.89,,1
LBT,MOT20-07,59.4544,31.16,30.5888,76.9196,40,5,2530,9868,70.1882,90.1797,19.4054,50.7441,20.0267,83.0687,56.1293,72.1165,80.4236,4.32479,1023,990,70.1882,90.1797,601,35.5976,27.7061,,1
LBT,MOT20-06,39.7757,15.9922,16.3841,75.522,34,94,5500,71104,46.4405,91.8097,7.96421,34.5218,8.08919,80.8352,36.621,72.3972,79.3033,5.45635,3348,3290,46.4405,91.8097,601,23.8039,12.0408,,1
LBT,MOT20-04,76.8407,39.2,38.0167,79.7125,353,38,6095,53284,80.5592,97.3137,23.5081,62.1709,24.1273,83.8192,65.7564,79.4322,82.3873,2.93029,4097,4019,80.5592,97.3137,601,43.2764,35.8255,,1
ITM,OVERALL,50.5502,48.611,39.557,78.5543,397,320,19495,233940,54.7877,93.5656,38.0893,41.3447,42.9951,67.4683,44.0845,75.2965,81.2062,4.35253,2431,4533,54.7877,93.5656,2500,65.814,38.5377,44.3712,0
ITM,MOT20-08,16.1814,22.6658,17.4023,70.9357,6,115,4910,59586,23.099,78.4725,18.6185,16.5577,19.8855,64.1885,17.6227,59.9026,74.7976,6.09181,450,880,23.099,78.4725,2500,49.8334,14.6688,19.4814,0
ITM,MOT20-07,51.367,48.8562,37.9242,78.2846,31,18,801,15071,54.4697,95.7464,35.711,40.8483,40.5117,64.3494,43.2999,76.1162,81.2507,1.36923,226,446,54.4697,95.7464,2500,67.3676,38.3251,4.1491,0
ITM,MOT20-06,24.7791,28.4773,22.1408,72.2208,15,139,7308,91688,30.9355,84.8936,22.2944,22.259,23.8865,65.3395,23.6488,64.9083,76.0647,7.25,865,1740,30.9355,84.8936,2500,53.3125,19.4272,27.9614,0
ITM,MOT20-04,72.6504,61.4152,49.6086,80.4979,345,48,6476,67595,75.3379,96.9591,42.6065,57.9072,48.2986,67.9179,61.5584,79.2315,82.7931,3.11346,890,1467,75.3379,96.9591,2500,70.228,54.5676,11.8135,0
HOMI_Tracker,OVERALL,51.2413,42.9925,37.2711,79.6102,423,312,16094,232259,55.1126,94.6578,32.894,42.4733,34.7791,78.8914,44.9607,77.2269,82.2309,3.59321,3937,6458,55.1126,94.6578,600,58.4168,34.012,71.4355,0
HOMI_Tracker,MOT20-08,16.9674,21.5457,18.0951,77.2605,11,113,3713,60174,22.3401,82.3384,19.3702,17.0241,20.2353,78.5912,18.0016,66.3544,80.0605,4.6067,450,798,22.3401,82.3384,600,50.478,13.6957,20.1432,0
HOMI_Tracker,MOT20-07,47.9412,44.8179,36.63,79.5609,24,20,1161,15772,52.3519,93.7209,33.6394,40.314,35.4238,80.675,42.8443,76.7211,82.3411,1.98462,299,419,52.3519,93.7209,600,62.5257,34.9264,5.71135,0
HOMI_Tracker,MOT20-06,23.9415,25.1947,21.0807,78.2224,15,152,4039,96080,27.6272,90.0801,20.9363,21.3308,21.956,79.6785,22.2841,72.6695,80.9438,4.00694,854,1576,27.6272,90.0801,600,53.6718,16.4609,30.9116,0
HOMI_Tracker,MOT20-04,74.5523,53.3051,46.2856,80.0424,373,27,7181,60233,78.0239,96.7512,35.9005,60.0141,37.9994,78.5989,63.8214,79.1426,82.625,3.4524,2334,3665,78.0239,96.7512,600,59.7022,48.1462,29.9139,0
GSDT_V2,OVERALL,67.1196,67.536,53.5935,79.0676,660,164,31507,135395,73.833,92.3811,52.7422,54.6725,58.5425,71.8386,59.8423,74.8757,81.7198,7.03438,3230,9878,73.833,92.3811,3000,76.0191,60.7561,,1
GSDT_V2,MOT20-08,39.5914,48.9126,38.9547,78.0163,43,62,9728,36406,53.0148,80.8527,41.2957,36.9595,46.8208,64.6659,42.5845,64.9454,80.758,12.0695,673,1815,53.0148,80.8527,3000,61.7545,40.4922,,1
GSDT_V2,MOT20-07,75.0279,68.058,55.6012,79.2345,71,2,1870,6116,81.5232,93.5193,51.6675,60.5924,57.2624,72.6571,66.3581,76.1227,82.0566,3.19658,280,608,81.5232,93.5193,3000,73.0653,63.6929,,1
GSDT_V2,MOT20-06,49.9333,51.9688,40.8757,78.1503,91,67,9796,55409,58.2628,88.7588,39.4108,42.5725,44.6873,62.9671,46.8385,71.3548,80.69,9.71825,1262,3566,58.2628,88.7588,3000,65.5696,43.041,,1
GSDT_V2,MOT20-04,82.2711,78.6341,62.0551,79.5309,455,33,10113,37464,86.3312,95.9012,59.0767,65.3605,65.0755,75.7415,70.2327,78.0182,82.1926,4.86202,1015,3889,86.3312,95.9012,3000,82.9925,74.7107,,1
GSDT,OVERALL,67.0575,67.5228,53.5935,79.0739,660,164,31913,135409,73.8303,92.2902,52.7422,54.6725,58.5425,71.8386,59.8423,74.8757,81.7198,7.12503,3131,9875,73.8303,92.2902,4800,75.9643,60.7698,42.4081,1
GSDT,MOT20-08,39.4146,48.9227,38.9547,78.0628,43,62,9916,36420,52.9967,80.5492,41.2957,36.9595,46.8208,64.6659,42.5845,64.9454,80.758,12.3027,608,1813,52.9967,80.5492,4800,61.6399,40.5555,11.4724,1
GSDT,MOT20-07,75.0249,68.0569,55.6012,79.2337,71,2,1870,6115,81.5262,93.5195,51.6675,60.5924,57.2624,72.6571,66.3581,76.1227,82.0566,3.19658,282,608,81.5262,93.5195,4800,73.0628,63.6929,3.45901,1
GSDT,MOT20-06,49.8301,51.9471,40.8757,78.1569,91,67,9966,55410,58.2621,88.5859,39.4108,42.5725,44.6873,62.9671,46.8385,71.3548,80.69,9.8869,1228,3564,58.2621,88.5859,4800,65.4656,43.0561,21.0772,1
GSDT,MOT20-04,82.2543,78.6277,62.0551,79.5309,455,33,10161,37464,86.3312,95.8826,59.0767,65.3605,65.0755,75.7415,70.2327,78.0182,82.1926,4.8851,1013,3890,86.3312,95.8826,4800,82.9772,74.7114,11.7339,1
GNNMatch,OVERALL,54.5498,49.0061,40.1757,79.4212,407,317,9522,223611,56.784,96.8609,37.0239,43.788,43.9862,58.7294,45.9457,78.3815,82.015,2.12592,2038,2456,56.784,96.8609,86400,66.2999,38.8678,35.8904,0
GNNMatch,MOT20-08,22.775,23.1994,20.7896,77.7282,14,113,2676,56788,26.71,88.5504,21.4864,20.2908,27.1468,43.0029,21.2815,70.5806,80.1442,3.3201,373,420,26.71,88.5504,86400,50.0556,15.0986,13.9648,0
GNNMatch,MOT20-07,52.3156,47.4552,38.5168,80.4899,31,20,460,15165,54.1857,97.4995,35.9647,41.6682,42.8527,59.5017,43.8459,78.8987,82.6955,0.786325,159,185,54.1857,97.4995,86400,66.4221,36.9143,2.93436,0
GNNMatch,MOT20-06,31.9524,29.0975,24.5355,77.8222,31,130,2666,86996,34.4697,94.4948,22.9097,26.3913,27.9688,43.3691,27.4501,75.2715,80.6889,2.64484,676,810,34.4697,94.4948,86400,54.4324,19.8558,19.6114,0
GNNMatch,MOT20-04,74.7479,61.9154,49.4653,79.8463,331,54,3720,64662,76.408,98.2547,41.5827,59.0534,49.0405,63.3701,62.1305,79.8984,82.4353,1.78846,830,1041,76.408,98.2547,86400,70.7669,55.032,10.8627,0
GMTrack,OVERALL,37.2863,37.8167,29.5929,77.758,101,425,12508,308305,40.4156,94.3563,28.5221,30.8864,33.409,56.915,32.2473,75.3035,80.3881,2.79259,3684,13943,40.4156,94.3563,400,63.0527,27.0073,91.1528,1
GMTrack,MOT20-08,29.6487,33.5835,26.8181,77.3487,12,99,450,53767,30.6089,98.138,30.7585,23.4635,35.6925,59.3296,24.0504,77.1165,80.5917,0.558313,294,1298,30.6089,98.138,400,70.6294,22.0291,9.60505,1
GMTrack,MOT20-07,55.0014,54.0348,41.4275,79.22,28,17,801,13878,58.0738,95.9998,38.4444,44.8902,44.2809,64.9573,47.2234,78.0634,81.6284,1.36923,216,593,58.0738,95.9998,400,71.679,43.3612,3.71941,1
GMTrack,MOT20-06,36.371,36.9867,29.8173,79.0882,31,123,949,82824,37.6123,98.1349,30.5798,29.1868,35.2881,57.5723,30.0766,78.4935,81.6986,0.941468,699,2675,37.6123,98.1349,400,66.7446,25.5813,18.5843,1
GMTrack,MOT20-04,37.7494,37.1206,28.4609,77.0284,30,186,10308,157836,42.4133,91.855,25.4162,32.0616,30.1548,54.633,33.8073,73.2378,79.5868,4.95577,2475,9377,42.4133,91.855,400,58.7566,27.1304,58.3544,1
GMPHD_Rd20,OVERALL,44.6518,43.5325,35.568,77.5347,293,274,42778,236116,54.3672,86.8005,32.0471,39.9294,34.1082,72.0107,44.0038,70.2691,80.2305,9.55079,7492,11153,54.3672,86.8005,177.92,56.5174,35.3995,137.804,0
GMPHD_Rd20,MOT20-08,11.4708,29.2163,23.3776,70.171,12,90,15069,52575,32.1473,62.3068,25.7718,21.6165,27.7282,64.4747,25.4252,49.3091,73.7977,18.696,952,1688,32.1473,62.3068,177.92,42.9211,22.1452,29.6137,0
GMPHD_Rd20,MOT20-07,50.1344,51.1938,39.9056,76.4937,37,19,1864,14204,57.0889,91.0216,39.0275,41.453,41.4917,75.687,45.0261,71.7993,79.8867,3.18632,438,578,57.0889,91.0216,177.92,66.4082,41.6513,7.67224,0
GMPHD_Rd20,MOT20-06,23.6281,34.3102,26.9468,72.0327,29,99,18448,81264,38.7874,73.6235,27.2816,27.0038,28.9524,69.4509,30.5922,58.0862,75.5966,18.3016,1677,2886,38.7874,73.6235,177.92,49.7176,26.193,43.2357,0
GMPHD_Rd20,MOT20-04,63.5531,50.2456,41.5068,80.1496,215,66,7397,88073,67.8664,96.1754,33.0261,52.5068,35.1453,72.3249,55.6287,78.8396,82.5314,3.55625,4425,6001,67.8664,96.1754,177.92,60.725,42.8507,65.2016,0
FGRNetIV,OVERALL,55.3855,52.6812,42.5407,79.4395,508,221,38973,189715,63.3349,89.3715,39.2541,46.3462,46.3812,62.2468,51.3792,72.5094,81.8225,8.70127,2159,3839,63.3349,89.3715,3500,63.5097,45.0074,34.0887,0
FGRNetIV,MOT20-08,28.1348,36.3593,29.6201,76.392,20,76,7414,47974,38.0853,79.9209,32.6179,27.1484,37.958,57.6259,30.1982,63.389,79.1493,9.19851,296,491,38.0853,79.9209,3500,56.3292,26.843,7.77203,0
FGRNetIV,MOT20-07,55.6841,51.7821,41.0259,80.0588,38,12,843,13652,58.7565,95.8456,37.9736,44.7306,43.1743,67.2608,47.5947,77.6381,82.5706,1.44103,174,256,58.7565,95.8456,3500,68.1254,41.7631,2.96137,0
FGRNetIV,MOT20-06,35.1778,37.3042,29.518,76.8503,46,90,11312,74050,44.2214,83.8444,27.8283,31.5431,32.5663,55.3496,34.9435,66.2704,79.4742,11.2222,694,1127,44.2214,83.8444,3500,54.0168,28.4896,15.6938,0
FGRNetIV,MOT20-04,72.8412,62.4828,50.4253,80.4842,404,43,19404,54039,80.2838,91.8964,43.1049,59.2024,51.1926,64.0535,65.785,75.3048,82.7503,9.32885,995,1965,80.2838,91.8964,3500,67.0017,58.535,12.3935,0
FBMOT,OVERALL,63.5098,65.1426,52.0496,77.3961,810,122,82835,102772,80.1378,83.3494,50.6718,53.8039,57.5575,68.2049,64.0628,66.7497,80.2836,18.4941,3203,5206,80.1378,83.3494,500,66.4479,63.8876,39.9686,1
FBMOT,MOT20-08,34.5233,50.4367,39.5312,75.6447,66,28,22123,27894,64.0003,69.1506,40.5118,38.849,47.2988,60.1725,50.5414,54.9342,78.539,27.4479,717,1249,64.0003,69.1506,500,52.4661,48.5584,11.2031,1
FBMOT,MOT20-07,73.4479,68.0963,54.9221,77.7795,81,4,3784,4749,85.653,88.225,51.4653,59.3572,57.9113,67.4334,68.5311,70.6022,80.7332,6.46838,256,301,85.653,88.225,500,69.1187,67.1037,2.9888,1
FBMOT,MOT20-06,43.6813,49.7757,39.9644,75.0816,123,64,28221,45144,65.995,75.6367,38.017,42.3724,43.6953,60.3301,51.9992,59.7731,78.1417,27.997,1402,2157,65.995,75.6367,500,53.4118,46.6032,21.244,1
FBMOT,MOT20-04,80.1083,75.6908,60.1729,78.5152,540,26,28707,24985,90.8842,89.6665,56.7988,63.9949,64.0768,72.3893,73.1889,72.2387,81.3578,13.8014,828,1499,90.8842,89.6665,500,75.1838,76.2047,9.1105,1
Fair,OVERALL,61.8141,67.2762,54.5724,78.6243,855,94,103440,88901,82.8186,80.5551,54.705,54.7333,60.6959,71.1041,67.3089,65.5812,81.143,23.0944,5243,7874,82.8186,80.5551,340.2,66.3568,68.2214,63.307,1
Fair,MOT20-08,27.0404,49.5389,41.209,77.2461,80,28,32104,23447,69.7396,62.7309,43.5899,39.2664,50.1723,62.9666,56.2165,50.7998,79.6979,39.8313,981,1739,69.7396,62.7309,340.2,47.0496,52.3063,14.0666,1
Fair,MOT20-07,75.5566,69.9503,56.6936,79.5897,85,1,2988,4770,85.5896,90.4595,52.9368,61.5064,58.8702,71.3002,69.6024,73.5721,81.8382,5.10769,333,445,85.5896,90.4595,340.2,71.9404,68.0674,3.89066,1
Fair,MOT20-06,38.5456,49.934,41.4017,76.6081,134,45,42384,36744,72.3224,69.3751,39.5164,43.7761,45.1334,60.4606,58.3103,56.1079,79.0954,42.0476,2457,3238,72.3224,69.3751,340.2,48.9165,50.9947,33.9729,1
Fair,MOT20-04,81.2554,80.7849,64.2753,79.5865,556,20,25964,23940,91.2654,90.5964,62.9467,65.8078,68.8806,76.6966,74.5264,74.0015,82.2005,12.4827,1472,2452,91.2654,90.5964,340.2,80.4888,81.0832,16.1288,1
D4C,OVERALL,54.7758,64.4195,51.4825,77.6931,713,109,121939,106271,79.4616,77.1262,51.5717,51.7335,57.587,70.6583,64.8473,62.9414,80.103,27.2246,5792,10119,79.4616,77.1262,819.6,63.4729,65.3949,,0
D4C,MOT20-08,35.1427,51.6423,41.9977,76.9442,71,25,25712,23313,69.9125,67.8129,42.8966,41.4404,48.1536,67.189,56.1589,54.4724,79.6239,31.9007,1229,2020,69.9125,67.8129,819.6,50.8669,52.4418,,0
D4C,MOT20-07,71.1127,69.3665,55.4596,78.6295,68,6,3104,6117,81.5202,89.6836,52.4501,59.1185,57.5799,72.6232,66.4025,73.052,81.0862,5.30598,341,622,81.5202,89.6836,819.6,72.8397,66.2095,,0
D4C,MOT20-06,41.5767,49.878,40.596,75.4592,125,41,36480,38162,71.2542,72.1686,37.7745,44.2555,42.4297,61.9017,56.9396,57.6703,78.2989,36.1905,2919,4067,71.2542,72.1686,819.6,50.198,49.562,,0
D4C,MOT20-04,64.7462,74.1958,58.2709,78.6558,449,37,56643,38679,85.8879,80.6049,58.7605,58.0039,65.5065,74.547,70.946,66.582,80.8219,27.2322,1303,3410,85.8879,80.6049,819.6,71.9139,76.6272,,0
CSTrack,OVERALL,66.5734,68.6148,53.9637,78.828,626,192,25404,144358,72.1007,93.6246,53.9865,54.1564,57.5624,77.8421,58.4684,75.9587,81.5387,5.6718,3196,7632,72.1007,93.6246,1000,78.8565,60.7277,44.3269,1
CSTrack,MOT20-08,42.5107,50.6507,39.2908,78.4729,34,71,3796,40037,48.3287,90.796,42.61,36.3563,45.5932,75.3525,39.0165,73.39,81.3245,4.70968,712,1501,48.3287,90.796,1000,72.9045,38.8054,14.7325,1
CSTrack,MOT20-07,73.4661,66.3693,55.0282,79.7766,73,3,2475,5910,82.1456,91.6571,50.0073,61.2582,54.2051,76.1202,67.8943,75.7685,82.4134,4.23077,398,523,82.1456,91.6571,1000,70.2117,62.9256,4.84506,1
CSTrack,MOT20-06,50.6738,53.8679,43.6781,80.1913,80,86,6234,58185,56.1718,92.2852,44.0627,43.4548,47.3218,75.1102,46.5231,76.5204,82.4295,6.18452,1065,2547,56.1718,92.2852,1000,71.1841,43.328,18.9597,1
CSTrack,MOT20-04,80.2447,79.0272,61.2304,78.3399,439,32,12899,40226,85.3235,94.7726,59.2831,63.4459,62.9471,78.972,68.6151,76.2268,81.1675,6.20144,1021,3061,85.3235,94.7726,1000,83.4031,75.0876,11.9662,1
center_reid,OVERALL,56.6098,64.9622,51.2209,77.0219,668,141,98181,121688,76.482,80.122,51.7777,50.9753,57.0142,72.2732,61.6408,64.6544,79.7377,21.9203,4643,9188,76.482,80.122,640,66.5081,63.4866,60.7071,0
center_reid,MOT20-08,29.6397,49.043,39.1979,76.2913,53,42,24136,29476,61.9586,66.5447,41.7087,37.0616,47.1836,66.2092,49.3147,53.0907,79.1589,29.9454,906,1970,61.9586,66.5447,640,50.858,47.353,14.6227,0
center_reid,MOT20-07,70.0795,66.2481,53.0127,77.3516,72,5,3589,5922,82.1093,88.3353,48.8312,58.0592,53.6266,71.8336,66.0467,71.0616,80.3239,6.13504,393,602,82.1093,88.3353,640,68.7598,63.9135,4.7863,0
center_reid,MOT20-06,39.5587,49.9156,39.9095,74.797,113,54,33287,44684,66.3415,72.5717,38.7949,41.522,42.6206,65.8799,52.4016,57.4708,77.9204,33.0228,2269,3676,66.3415,72.5717,640,52.2594,47.773,34.2018,0
center_reid,MOT20-04,70.8666,76.2241,59.1888,77.9771,430,40,37169,41606,84.82,86.2157,58.8905,59.7204,64.587,75.8103,69.0685,70.2373,80.4797,17.8697,1075,2940,84.82,86.2157,640,76.8512,75.6071,12.6739,0
BBT,OVERALL,46.8388,42.1509,35.7651,77.9982,312,289,35014,236176,54.3556,88.9289,32.2326,40.1578,35.0521,70.1808,43.8492,71.7493,80.6265,7.81737,3880,7207,54.3556,88.9289,560,55.5561,33.9573,71.3818,0
BBT,MOT20-08,12.6039,23.7792,20.2093,70.5459,14,93,13716,53331,31.1716,63.7804,19.882,21.0627,22.0867,57.3845,24.5117,50.1735,74.0694,17.0174,671,1421,31.1716,63.7804,560,36.217,17.7004,21.526,0
BBT,MOT20-07,52.1283,47.1703,37.8438,76.8417,36,19,1352,14245,56.965,93.3096,35.0243,41.6112,38.5165,70.3118,44.7425,73.2925,80.2164,2.31111,249,382,56.965,93.3096,560,62.2179,37.9837,4.3711,0
BBT,MOT20-06,25.054,28.2688,23.5589,72.3787,31,107,15695,82573,37.8014,76.176,21.483,26.2818,23.5435,61.7772,29.4259,59.3098,75.8689,15.5704,1228,2295,37.8014,76.176,560,42.6175,21.1484,32.4856,0
BBT,MOT20-04,66.43,52.0335,43.6774,80.5708,231,70,4251,86027,68.6129,97.7895,35.8965,53.4221,38.8319,73.1032,56.1941,80.0952,82.8758,2.04375,1732,3109,68.6129,97.7895,560,63.0967,44.2711,25.2431,0
ApLift,OVERALL,58.8896,56.4849,46.6146,79.5893,513,264,17739,192736,62.751,94.8197,45.2281,48.2459,48.1469,76.7792,51.2973,77.5126,82.1505,3.96048,2241,2112,62.751,94.8197,9999,70.9181,46.9331,,0
ApLift,MOT20-08,26.4816,33.8377,27.5797,77.3829,20,98,3702,52924,31.6969,86.9011,31.8348,24.064,34.26,69.9968,25.5346,70.0066,80.0756,4.59305,339,333,31.6969,86.9011,9999,63.3041,23.0899,,0
ApLift,MOT20-07,56.9046,54.7229,44.0464,79.5324,40,15,936,13135,60.3184,95.522,42.2137,46.4185,45.3183,75.852,49.2313,77.9641,82.237,1.6,194,195,60.3184,95.522,9999,70.6918,44.6391,,0
ApLift,MOT20-06,36.0945,36.7543,30.1005,77.7757,41,111,4786,79313,40.257,91.7809,29.5598,30.8428,31.4793,71.4129,32.5053,74.1078,80.649,4.74802,740,744,40.257,91.7809,9999,60.2748,26.4378,,0
ApLift,MOT20-04,79.3322,68.7772,56.6164,80.2608,412,40,8315,47364,82.7192,96.4622,50.5213,63.6551,53.6641,78.7668,67.9322,79.2186,82.73,3.9976,968,840,82.7192,96.4622,9999,74.4906,63.8779,,0
ALBOD,OVERALL,56.4821,51.1433,43.4616,79.4218,506,228,36864,184582,64.3269,90.0289,39.6679,48.0006,42.6119,76.5023,52.7664,73.9031,81.9621,8.23041,3727,4308,64.3269,90.0289,3600,61.3606,43.843,57.9385,0
ALBOD,MOT20-08,25.0297,36.1321,28.8209,74.3176,24,78,11486,46151,40.438,73.1755,30.4973,27.3999,33.6107,66.8514,31.7354,57.543,77.6168,14.2506,453,732,40.438,73.1755,3600,50.7578,28.0497,11.2023,0
ALBOD,MOT20-07,53.802,50.8422,41.9138,80.0124,36,13,1945,13114,60.3819,91.1317,38.8074,45.6827,42.6084,70.7286,49.7444,75.077,82.5228,3.32479,233,319,60.3819,91.1317,3600,63.7881,42.2646,3.85877,0
ALBOD,MOT20-06,31.4093,36.4588,28.7052,73.5771,43,94,16835,73176,44.8797,77.9693,26.7648,31.0526,29.6065,65.0145,35.1945,61.261,77.0585,16.7014,1048,1559,44.8797,77.9693,3600,49.8992,28.7224,23.3513,0
ALBOD,MOT20-04,77.8418,60.8804,52.7374,81.6581,403,43,6598,52141,80.9763,97.113,43.8269,63.7331,46.5994,80.4335,67.5881,81.0696,83.8622,3.17212,1993,1698,80.9763,97.113,3600,66.9464,55.8223,24.6121,0
AGNNTrack,OVERALL,70.7674,66.1205,56.3981,79.8131,830,135,47397,101754,80.3346,89.7646,54.4622,58.6617,59.5162,74.7904,66.029,73.8731,82.3381,10.582,2106,2678,80.3346,89.7646,265.3,70.0012,62.6474,26.2154,1
AGNNTrack,MOT20-08,42.3662,47.8742,39.9674,77.1281,50,52,11371,32814,57.6506,79.7095,40.5076,39.7539,44.1905,69.9535,46.3017,64.2924,79.8055,14.1079,472,611,57.6506,79.7095,265.3,57.0332,41.2498,8.18725,1
AGNNTrack,MOT20-07,81.1758,65.766,57.2379,81.1677,87,2,1454,4590,86.1333,95.1477,49.4915,66.6222,54.9082,71.8793,72.0339,79.5806,83.5611,2.48547,187,198,86.1333,95.1477,265.3,69.2074,62.6507,2.17105,1
AGNNTrack,MOT20-06,54.7549,48.6465,42.5004,77.8121,117,58,14420,44723,66.3121,85.9254,38.394,47.3914,41.7672,68.8799,53.6839,69.735,80.6703,14.3056,923,1105,66.3121,85.9254,265.3,55.8407,43.0945,13.919,1
AGNNTrack,MOT20-04,85.2954,78.0921,65.7174,80.825,576,23,20152,19627,92.8391,92.6616,62.7775,68.9642,68.5095,77.7496,76.8603,76.7382,83.2296,9.68846,524,764,92.8391,92.6616,265.3,78.0175,78.1669,5.64418,1
\end{filecontents*}
\csvreader[
	respect underscore=true,
	head to column names,
	filter expr=	{	
		test{\ifcsvstrcmp{\csvcoli}{MOTer}} and test{\ifcsvstrcmp{\sequence}{OVERALL}} or
		test{\ifcsvstrcmp{\csvcoli}{TransCtr}} and test{\ifcsvstrcmp{\sequence}{OVERALL}} or
		test{\ifcsvstrcmp{\csvcoli}{GNNMatch}} and test{\ifcsvstrcmp{\sequence}{OVERALL}} or
		test{\ifcsvstrcmp{\csvcoli}{UnsupTrack}} and test{\ifcsvstrcmp{\sequence}{OVERALL}} or
		test{\ifcsvstrcmp{\csvcoli}{Tracktor++v2}} and test{\ifcsvstrcmp{\sequence}{OVERALL}} or
		test{\ifcsvstrcmp{\csvcoli}{center-reid}} and test{\ifcsvstrcmp{\sequence}{OVERALL}} or
		test{\ifcsvstrcmp{\csvcoli}{SFS}} and test{\ifcsvstrcmp{\sequence}{OVERALL}} or
		test{\ifcsvstrcmp{\csvcoli}{RTv1}} and test{\ifcsvstrcmp{\sequence}{OVERALL}} or
		test{\ifcsvstrcmp{\csvcoli}{ALBOD}} and test{\ifcsvstrcmp{\sequence}{OVERALL}} or
		test{\ifcsvstrcmp{\csvcoli}{FGRNetIV}} and test{\ifcsvstrcmp{\sequence}{OVERALL}} or
		test{\ifcsvstrcmp{\csvcoli}{D4C}} and test{\ifcsvstrcmp{\sequence}{OVERALL}} or
		test{\ifcsvstrcmp{\csvcoli}{GNNMatch}} and test{\ifcsvstrcmp{\sequence}{OVERALL}} or
		test{\ifcsvstrcmp{\csvcoli}{HOMI_Tracker}} and test{\ifcsvstrcmp{\sequence}{OVERALL}} or
		test{\ifcsvstrcmp{\csvcoli}{ITM}} and test{\ifcsvstrcmp{\sequence}{OVERALL}}
	}
	]
	{MOT20_results.csv}{}
	{\\\hline
		\csvcoli
		&\ifcsvstrcmp{\csvcoli}{RTv1}{\bfseries}{}\num\HOTA
		&\ifcsvstrcmp{\csvcoli}{MOTer}{\bfseries}{}\num\MOTA
		&\ifcsvstrcmp{\csvcoli}{RTv1}{\bfseries}{}\num\csvcoliv
		&\ifcsvstrcmp{\csvcoli}{UnsupTrack}{\bfseries}{}\num\MOTP
		&\ifcsvstrcmp{\csvcoli}{HOMI_Tracker}{\bfseries}{}\num\runtime
	} \\\hline\hline
	\csvreader[
	respect underscore=true,
	head to column names,
	
	filter expr=	{	
		test{\ifcsvstrcmp{\csvcoli}{SORT20}} and test{\ifcsvstrcmp{\sequence}{OVERALL}} or
		test{\ifcsvstrcmp{\csvcoli}{Surveily}} and test{\ifcsvstrcmp{\sequence}{OVERALL}} or
		test{\ifcsvstrcmp{\csvcoli}{GMPHD_Rd20}} and test{\ifcsvstrcmp{\sequence}{OVERALL}}
	}
	]
	{MOT20_results.csv}{}
	{\csvcoli
		&\num\HOTA
		&\num\MOTA
		&\num\csvcoliv
		&\ifcsvstrcmp{\csvcoli}{SORT20}{\bfseries}{}\num\MOTP
		&\ifcsvstrcmp{\csvcoli}{SORT20}{\bfseries}{}\num\runtime
		\\\hline}
	LTSiam & \textbf{40.4} & \textbf{46.5} & \textbf{49.4} & 77.1 & 148\\
	\hline
\end{tabular}
\end{center}
\caption{Results on the MOT20 benchmark for online algorithms, devided into two parts for real-time solutions (bottom) and non-real-time (top). Here, the first four columns of the MOT20 benchmark results are shown. The full list is available at \url{motchallenge.net/results/MOT20}. The column RT shows the runtime of the corresponding algorithm for all 4479 frames of the test scenes.}
\label{tab:res}
\end{table}
\catcode`\_=3



As motivated in the introduction, we evaluate the effectiveness and real-time capability based on the MOT20 benchmark~\cite{MOTChallenge20}. The tracking results of the test scenes must be submitted, ground truth data for own evaluation is not provided. Results are then automatically generated, listed in Table~\ref{tab:res}.

Note that in contrast to all other values the runtime is provided by the authors of the algorithms. 
Our evaluation system is equipped with an Intel Xeon Platinum 8180 Processor. We did not parallelize the algorithm, so only a single core is utilized except for the GPU parts. Running on the GPU is the inference of a fingerprint and the cosine similarity (in different steps). For this, we utilize an NVIDIA V100 GPU.

As can be seen in Table~\ref{tab:res} and Fig.~\ref{fig:hota_cmp}, among real-time capable approaches our proposed method performs best in HOTA, MOTA and IDF1 and even outperforms non-real-time capable approaches.

As described in section~\ref{sec:rel}, SORT20 follows a similar approach ignoring appearance features. Thus, the proposed LTSiam model can be assumed as the main cause for the improved performance. GMPHD\_Rd20 applies a fast CNN called IdNet to include appearance features. However, caused by the design where training differs from inference, this leads to inferior results.

\section{Conclusion and Outlook}

In this paper, we present a novel approach for real-time capable multi-object tracking in a surveillance scenario. 
It is based on the basic idea of associating given detections with tracks. 
For this, we utilize the Hungarian algorithm minimizing a cost matrix. Furthermore, we introduce LTSiam to derive fingerprints to improve results in linear time. The evaluation shows that this outperforms other real-time capable approaches.

In future research, utilizing fingerprints could help to distinguish between different reasons for the disappearance of a person. To be precise, case 3 in Sec.~\ref{sec:dis} could be identified by comparing the fingerprint of the patch at the current tracking position with the last patch where the person is known to be visible. However, as the current tracking position is required and vice versa, the fingerprint must be inferred in each frame.
Further research is required to avoid the additional overhead.

\section*{Acknowledgments}
This research has been funded by the Federal Ministry of Education and Research of Germany as part of the competence center for machine learning ML2R (01IS18038B).

{\small
\bibliographystyle{ieee_fullname}
\bibliography{literature}

\begin{thebibliography}{10}\itemsep=-1pt

\bibitem{baisa2020occlusionrobust}
Nathanael~L. Baisa.
\newblock Occlusion-robust online multi-object visual tracking using a gm-phd
  filter with cnn-based re-identification.
\newblock 2020.
\newblock arXiv: 1912.05949.

\bibitem{bergmann2019tracking}
Philipp Bergmann, Tim Meinhardt, and Laura Leal-Taixe.
\newblock Tracking without bells and whistles, 2019.
\newblock arXiv: 1903.05625.

\bibitem{bewley2016simple}
Alex Bewley, Zongyuan Ge, Lionel Ott, Fabio Ramos, and Ben Upcroft.
\newblock Simple online and realtime tracking.
\newblock In {\em 2016 IEEE international conference on image processing
  (ICIP)}, pages 3464--3468. IEEE, 2016.

\bibitem{MOTChallenge20}
P. Dendorfer, H. Rezatofighi, A. Milan, J. Shi, D. Cremers, I. Reid, S. Roth,
  K. Schindler, and L. Leal-Taix\'{e}.
\newblock Mot20: A benchmark for multi object tracking in crowded scenes.
\newblock Mar. 2020.
\newblock arXiv: 2003.09003.

\bibitem{ding2015deep}
Shengyong Ding, Liang Lin, Guangrun Wang, and Hongyang Chao.
\newblock Deep feature learning with relative distance comparison for person
  re-identification.
\newblock {\em Pattern Recognition}, 48(10):2993--3003, 2015.

\bibitem{karthik2020simple}
Shyamgopal Karthik, Ameya Prabhu, and Vineet Gandhi.
\newblock Simple unsupervised multi-object tracking, 2020.
\newblock arXiv: 2006.02609.

\bibitem{Luiten_2020}
Jonathon Luiten, Aljossa Ossep, Patrick Dendorfer, Philip Torr, Andreas Geiger,
  Laura Leal-Taixe, and Bastian Leibe.
\newblock Hota: A higher order metric for evaluating multi-object tracking.
\newblock {\em International Journal of Computer Vision}, 129(2):548–578, Oct
  2020.

\bibitem{papakis2021gcnnmatch}
Ioannis Papakis, Abhijit Sarkar, and Anuj Karpatne.
\newblock Gcnnmatch: Graph convolutional neural networks for multi-object
  tracking via sinkhorn normalization, 2021.

\bibitem{pflugfelder2017depth}
Roman Pflugfelder.
\newblock An in-depth analysis of visual tracking with siamese neural networks.
\newblock {\em arXiv preprint arXiv:1707.00569}, 2017.

\bibitem{conf/nips/RenHGS15}
Shaoqing Ren, Kaiming He, Ross~B. Girshick, and Jian Sun.
\newblock Faster r-cnn: Towards real-time object detection with region proposal
  networks.
\newblock In Corinna Cortes, Neil~D. Lawrence, Daniel~D. Lee, Masashi Sugiyama,
  and Roman Garnett, editors, {\em NIPS}, pages 91--99, 2015.

\bibitem{varior2016gated}
Rahul~Rama Varior, Mrinal Haloi, and Gang Wang.
\newblock Gated siamese convolutional neural network architecture for human
  re-identification.
\newblock In {\em European conference on computer vision}, pages 791--808.
  Springer, 2016.

\bibitem{xu2021transcenter}
Yihong Xu, Yutong Ban, Guillaume Delorme, Chuang Gan, Daniela Rus, and Xavier
  Alameda-Pineda.
\newblock Transcenter: Transformers with dense queries for multiple-object
  tracking, 2021.
\newblock arXiv: 2103.15145.

\end{thebibliography}
}

\end{document}